# "Understanding AI": Semantic Grounding in Large Language Models

Holger Lyre

Chair for Theoretical Philosophy & Center for Behavioral Brain Sciences,
University of Magdeburg, Magdeburg, Germany

February 2024

**Abstract:** Do LLMs understand the meaning of the texts they generate? Do they possess a semantic grounding? And how could we understand whether and what they understand? I start the paper with the observation that we have recently witnessed a generative turn in AI, since generative models, including LLMs, are key for self-supervised learning. To assess the question of semantic grounding, I distinguish and discuss five methodological ways. The most promising way is to apply core assumptions of theories of meaning in philosophy of mind and language to LLMs. Grounding proves to be a gradual affair with a three-dimensional distinction between functional, social and causal grounding. LLMs show basic evidence in all three dimensions. A strong argument is that LLMs develop world models. Hence, LLMs are neither stochastic parrots nor semantic zombies, but already understand the language they generate, at least in an elementary sense.

Large language models (LLMs) have taken the world by storm. With the advent of these amazing systems, we have truly entered a new AI era. LLMs are capable of generating any conceivable type of text, fictional or non-fictional, whether for purposes of information or knowledge transfer, conversation, strategic planning, or program code generation. Indeed, the fact that we have "talking machines" is absolutely stunning. But do LLMs talk – and maybe even think – meaningfully? This question currently divides the experts. Some believe that LLMs like LaMDA, LLaMA and GPT are merely clever forms of autocompletion based on learned word statistics. From this point of view, LLMs are nothing but "stochastic parrots" (Bender et al. 2021). I will argue instead that there is every indication that LLMs already have an elementary semantic grounding and that they operate in a rudimentary meaningful way. In other words: LLMs already understand the language they generate, at least in an elementary sense.

Clarifying the question of semantic grounding is of high scientific and philosophical relevance. It is, at the same time, more than just an academic exercise. In order to properly assess the – positive and negative – potentials of modern generative AI, an evaluation of the semantic capabilities is of utmost importance, since machines acting in a genuinely meaningful way must be assumed incomparably more powerful than semantically "empty", stochastic parrots.

Again, the question of semantic grounding divides the experts. On the side of the skeptics,

Melanie Mitchell (2019, 2020) diagnoses a "Meaning Barrier": a lack of understanding, since AI systems "do not capture the rich meanings that humans bring to bear in perception, language, and reasoning" (Mitchell 2020). And in line with the stochastic parrot objection, Bender & Koller (2020) presented their octopus thought experiment, essentially a variant of Searle's Chinese room.[1] Here's the setting: Two human speakers, independently stranded on two desert islands, can communicate with each other via an underwater cable. A hyper-intelligent octopus, not familiar with the surface world, taps into the cable, listens in on the human conversation and learns to predict the speakers' outcomes with great accuracy. But no matter how accurate the prediction (and the possible imitation of one of the humans), the octopus will never be able to grasp the meaning of human language, Bender & Koller claim.

The core intuition behind the parrot and octopus claims is that lack of causal world contact amounts to lack of meaning in terms of reference. However, semantics encompasses more than just referential semantics. There is also more to learning word statistics, because languages and large amounts of text data somehow mirror the world. This brings us to the other, far less skeptical group of authors. More positive views on the prospects of semantic grounding were expressed, for example, by Piantadosi & Hill (2022), Søgaard (2022, 2023), Chalmers (2023), Coelho Mollo & Millière (2023), Piantadosi (2023), and Pavlick (2023). This paper sides with the more positive attitude of these authors.

The title of the paper contains a deliberate ambiguity: Do LLMs understand, and how do we understand whether and what they understand? The second question is explicitly taken into account here, since the starting point of the paper lies in methodological considerations. What general ways are there to assess the question of semantic grounding? I will argue in favor of a five-fold methodology, distinguishing extrospective from introspective as well as behavioral from non-behavioral methods. On the face of it, there are two ways to judge the grounding question. We may first judge it from the overall observable output and behavior of AI systems. This is, of course, the rationale behind the Turing test. It is, however, rather well-known that this is a limited method. The main problem is that one and the same behavior can in principle be brought about in quite different ways.

This motivates another method: looking under the hood of systems and exploring their internal mechanisms and functions. But in the case of deep learning neural networks, the notorious black box problem looms, i.e. the fact that sophisticated AI systems are notoriously opaque. This is indeed a problem that should not be underestimated. I would like to give it the following twist here: By bringing their rather simplistic and premature analogies of octopuses and parrots to LLMs, the semantic skeptics demonstrate their perhaps overly hasty assumptions about the inner workings of transformers – an instance of the black box problem.

To avoid falling into the same trap, I propose a method for assessing the question of grounding that consists in the top-down application of theories of meaning from the analytic philosophy of mind and language. This goes hand in hand with important insights into the nature of semantic grounding. First, it is by no means a simple yes-no question. Second, it is multi-criterial, and I will distinguish three basic types of grounding under the titles functional, social and causal grounding. Third, grounding is also gradual, in other words, it comes in degrees. My analysis aims to show that modern LLMs possess a decent functional, a weak social, and an indirect causal grounding. The strongest argument for grounding is that LLMs develop world models.

---

[1] See Lyre (2020, sec. 4.3) for a critical disussion of Searle's Chinese room.

The paper proceeds as follows: In the first section, I provide some philosophical background, introduce the notion of the *generative turn in AI* along with some technical background and will then present the methodology used in this paper. Section 2 discusses (and finally dismisses) behavioral and introspective methods. In the central section 3, the grounding question will be tackled by applying core assumptions of theories of meaning to LLMs under the headings functional, social, and causal grounding. I finish in section 4 with some concluding remarks on where the LLM world models reside and what the prospects of more direct forms of causal grounding are.

## 1 Introduction

In this introductory section, I will first provide some philosophical background and then explain the notion of the *generative turn in AI* together with some technical remarks on generative AI systems in general and transformers in particular. Finally, I will present the explicit methodology used in this paper.

### 1.1 Philosophical context

In philosophy of mind, it is standard to distinguish between two types of mental states: phenomenal states and representational states (or intentional states). Phenomenality aims at phenomenal consciousness: the what-it's-like-to-be in a mental state, its phenomenal character or quale. Phenomenal states are therefore individuated by qualia. Intentionality aims at the idea that mental states refer, represent and possess semantic content. Representational states are semantically individuated. Only this kind of states is of interest to us. In fact, there is little to suggest that phenomenal consciousness plays a role in semantic grounding. One empirical indication of this is provided by evidence from semantic priming experiments, which suggest the possibility of unconscious semantic processing (cf. Rohaut & Naccache 2018). In any case: questions about sentience or consciousness will not be considered here.

We may nevertheless borrow from the consciousness debate. Qualia enthusiasts claim that phenomenal character plays no functional or behavioral role. In other words, phenomenal zombies are conceivable. The science fiction genre knows such phenomenally empty characters as for instance the android "Commander Data" from "Star Trek: The Next Generation". By analogy, we can ask whether there are beings or systems that behave as if they possess representational states, but are in fact "semantically empty." This would amount to "semantic zombies." There are prominent examples of this in the analytic philosophy of mind (though not under this phrase): Block's "Blockhead", Searle's "Chinese Room", and Davidson's "Swampman". Thematically related is the fact that forms of mental externalism lead to causally inert semantic content, so-called wide content. The stochastic parrot thesis is just the special case of the semantic zombie claim applied to LLMs.

To deny semantic grounding of LLMs is to claim that LLMs are semantic zombies: they might behave like semantically grounded systems, but do in fact possess no semantic capabilities. The proponent of the zombie or stochastic parrot claim bears the burden of proof to show how it is possible that an AI system exhibits meaningful-like behavior despite being semantically empty. In the case of phenomenal zombies, this is accomplished by the idea that qualia are functionally inert. This path is, however, not open to the proponent of the semantic zombie claim. Mental representations with semantic content are typically postulated for the very reason of explaining behavior. In the context of LLMs, proponents of the stochastic parrot claim point to behavioral deviations of LLMs relative to human speakers, primarily the widespread tendency of such systems to confabulate (see section 2.2). I will, however, ultimately dismiss behavioral methods and prefer a top-down application of theories of semantics to evaluate semantic grounding (section 3).

**1.2 The generative turn in AI and the pre-history of Transformers**

In the broadest sense, generative AI refers to any kind of data-producing AI. In a narrower sense, it refers to *generative models*. Such models are distinguished from discriminative models in machine learning and are used to generate specific types of data based on a previously learned probability model of typical data sets. But generative models are also of fundamental importance in a further sense, as they provide the key concept for *self-supervised learning*.

The crucial point can be well illustrated by the autoencoder architecture (cf. Hinton & Salakhutdinov 2006). It consists of an encoder, a bottleneck, and a decoder. In the encoder part, a discriminative model (e.g. a classifier network) learns to extract the essential features and structures from datasets, resulting in a reduced data representation due to the bottleneck. In the decoder part, the data will be reconstructed from the reduced, latent representation. In principle, the system can now self-supervise whether learning was successful. This deep connection between generative models and the possibility of self-supervised learning paves the way to what I like to highlight as the *generative turn*. In their widely regarded Turing award paper, Bengio, LeCun and Hinton describe generative self-learning as "*a form of prediction or reconstruction called self-supervised learning which is training to "fill in the blanks" by predicting masked or corrupted portions of the data*." (Bengio et al. 2021, 62)

The core idea of generative AI can be traced back at least to the work of Geoffrey Hinton et al. on Boltzmann and Helmholtz machines (Ackley et al. 1985, Dayan et al. 1995).

Boltzmann machines (BM) are Hopfield networks with stochastic neurons that follow a Boltzmann probability distribution. In a BM, visible (V) and hidden (H) neurons are distinguished. This allows two phases of work: In the input phase, the V neurons are "clamped" to the environment and receive external input; the training data are sets of binary vectors over V. In the free-running phase (the V are now unclamped), the H-units are to model the V distributions based on features (i.e. latent variables). In its two phases of work, the BM contrasts the statistics of network activations on input versus free-running until a "thermal" equilibrium is reached. It proves beneficial to remove lateral connections (within layers), which leads to Restricted Boltzmann Machines (RBMs). These systems can be connected in series, resulting in Stacked RBMs and Deep Belief Nets (cf. Hinton 2014). They are among the basic architectures of Deep Learning.

The two most important developments of generative AI in the 2010s are GANs (generative adversarial networks) and Transformer architectures. GANs (Goodfellow et al. 2014) combine two competing networks: a generator, such as a Convolutional Neural Network, that creates content based on image input, and a discriminator that is trained to distinguish the generator data from real data. The goal of the generator is to produce data that the discriminator cannot distinguish from real data. Very well known is the creation of photorealistic images with GANs.

Transformer architectures (Vaswani et al. 2017) essentially consist of stacks of encoders and stacks of decoders. The input, a written sentence, is transformed into a vector representation via embedding, involving both position encoding and a multi-head attention mechanism. The latter is the architectural centerpiece, as it accounts for the crucial contextual correlations in the input. Modern LLMs are typically based on transformers.[2] Many of them, as for instance GPT-4, consist of decoders only (and can be seen as autoregressive models). LLM training is furthermore based on a two-step process. In the first step, the system is trained using huge amounts of data (including, for instance, the common crawl data corpus and the Wikipedia) on the basis of generative self-learning. This results in a foundation model that can now be adapted to a variety of downstream tasks in a second step. The fine-tuning of this second step is done by Reinforcement Learning from Human Feedback (RLHF).

### 1.3 Methodology: How to test semantics?

How could we possibly test whether a system or agent acts in a semantically meaningful way? A systematic approach, I claim, leads to a fivefold methodological distinction. First, extrospective and introspective methods must be distinguished, i.e. methods from the standpoint of an external, typically third-person perspective versus methods based on the first-person or self-perspective. In both cases, a behavioral method focusing on external performance can be characterized. The standard paradigm for the extrospective-behavioral method is the Turing test. Here, the AI system's semantic capabilities are evaluated in an operationalizable way based on performance equivalence with humans. In the case of extrospection, two other methods should be added. They can be characterized as top-down and bottom-up. The top-down method consists of applying theories, models, and explanations of semantics. This includes the common canon of theories of semantics and meaning as well as programs for naturalizing mental representation in analytic philosophy of mind and philosophy of language. In contrast, the bottom-up method focuses on the AI system's internal realization by looking under the hood of its inner architecture, functioning, and mechanisms for producing semantic capabilities.

---

[2] See Millière & Buckner (2024) for a recent comprehensive philosophical introduction to LLMs. See also Søgaard (2022) for a basic introduction in the context of language processing that simultaneously uncovers numerous errors in Landgrebe & Smith (2021).

While the extrospective approach comprises three methodological options, only two introspective methods can be distinguished. The introspective-behavioral method aims at self-reports. In some sense, they are the counterpart of Turing tests. The counterpart of the theoretical and explanatory top-down method are self-explanations and self-reports where a system reflects or justifies its own performance.

Hence, from a systematic point of view, we can distinguish five methods by which semantics can in principle be tested:

| | |
|---|---|
| E. Extrospective methods | (1) Behavioral: general performance |
| | (2) Top-down: application of theories of semantics |
| | (3) Bottom-up: mechanistic realization |
| I. Introspective methods | (1) Self-reports |
| | (2) Self-explanation |

I will turn to each of these methods below. While no method provides watertight results, the different methods are not equivalent in strength. Methods E1 and I1, as behavioral methods in general, turn out as empirically underdetermined. This is where the problem of semantic zombies becomes most visible (see sections 2.1 and 2.2). Method I2 has to presuppose that self-explanations are trustworthy, which is also fraught with problems (see section 2.2). And while it certainly makes sense to look under the hood of AI systems, method E3 eventually faces the black box problem. It has (mis)led some authors to the stochastic parrots thesis, as already pointed out in the introduction. What essentially remains is method E2. Although an obvious and methodologically sound choice, it has not yet received the attention it deserves in the literature.[3] Method E2 will therefore be central to section 3 and to the paper in general.

As already mentioned in the beginning, one important finding of the paper will be that semantic grounding cannot be assessed using a single criterion, but that grounding has at least three dimensions. Furthermore, semantic grounding isn't on/off but gradual. AI systems can be more or less grounded. This should be familiar: humans have more or less semantic capabilities depending on their living conditions. In newborns or patients with severe dementia, for example, grounding is greatly reduced to non-existent. And conversely, it is perfectly conceivable that the semantic groundling of more advanced AI systems may exceed that of our species.

---
[3] One notable exception is Coelho Mollo & Millière (2023).

## 2 Behavioral and introspective methods

### 2.1 The extrospective-behavioral method E1

What is achieved by modern LLMs? How do they perform? These are the central questions of extrospective and behavioral methods. In this spirit, Bubeck et al. (2023) conducted an extensive and widely regarded study testing the power spectrum and strengths, but also the limitations of GPT-4 (prior to its release in March 2023). The authors tested a range of applications: multimodal imaging, coding, mathematical skills, discriminative skills, and human interactions. They believe they can detect "sparks of AGI" in GPT-4, because some of the progress over GPT-3.5 is significant.

In many of their tests Bubeck et al. used combinations of domains that are almost certainly not included in the training data. For instance, an answer to the question "Can you write a proof that there are infinitely many primes, with every line that rhymes?" requires a creative combination of mathematical thinking, poetic expression, and natural language. And indeed, GPT-4 not only performs amazingly on this task, it can also masterfully rewrite the answer into another poetic style, such as Shakespeare's. The authors also tested the (non-modal) GPT-4 version for its ability to generate source code to produce images, for example by using the prompt "Draw a unicorn in TikZ." Here again, GPT-4 proves far more creative and variable than GPT-3.5, which the authors believe shows that the system develops visual ideas based solely on natural language descriptions. GPT-4 also demonstrates remarkable general understanding in human interaction. The fact that both GPT-4 and already GPT-3.5 have solid Theory-of-Mind-capabilities deserves special attention. Bubeck et al. tested several variations of the Sally-Anne test, i.e. the classical false belief design. They also tested the understanding of emotional or intentional states of characters in invented scenarios, some of which showed strange behaviors to complicate matters. Overall, GPT-4 provided more sophisticated and far-sighted answers than GPT-3.5.

A further, truly remarkable example of AI performance comes from the field of brain and mind reading: With the help of an LLM, Tang et al. (2023) were recently able to decode human thoughts in a rudimentary form based on fMRI data. In the encoding part of the experiment, the researchers recorded fMRI data from subjects while they listened to podcasts. With this, they trained an LLM on predictions about how subjects' brains respond to different words and word constellations. In the decoding part of the experiment, fMRI data were recorded while the subjects watched a movie without sound, for example. The LLM was now able to generate sentences, some of which described quite well, if only elementarily, what people saw in the movies.

The list of achievements of modern LLMs could easily be continued, but how does all this relate to the otherwise also well-known fatal errors of LLMs, in particular their tendency to "hallucinate"?[4] LLMs regularly invent data and false answers, not infrequently in the mode of unbreakable conviction. First of all, two things can be countered against this: Unlike humans, LLMs learn from mind-bogglingly large amounts of data. At the same time, and unlike the typical human learning environment, such data are riddled with inconsistencies and contradictions. Given this, the extensive consistency of LLMs is nothing short of remarkable. Second, it is no secret that humans also invent information, not infrequently intentionally. A common cause of human confabulations are cognitive biases. We also touch upon this in the following section 2.2.

In sum, from an extrospective-behavioral perspective, LLMs show impressive and at times breathtaking abilities, but they also come with some perplexing weaknesses. But even if the positive side prevails, purely behavior-based methods are limited in principle, as is known from the long-standing debate about the Turing test. The main objection is that one and the same behavior can in principle be brought about in quite different ways. Mere "brute force" mechanisms, in particular, should not be considered intelligent. A classic example of this is "blockhead", a hypothetical AI system capable to conduct an outwardly meaningful conversation by simply following a pre-programmed look-up table. According to Block (1981), such a system is evidently a semantic zombie. But at least two things can be said against it. First, as McDermott (2014) has shown, the combinatorial tree behind such a look-up table explodes so rapidly that a human-like blockhead is not a nomological, but at best a mere logical possibility. And second, of course, LLM transformer architectures do not correspond to simple look-up tables. Nevertheless, the blockhead objection demonstrates that the problem of semantic zombies can never be fully and fundamentally overcome by E1 methods.

**2.2 Introspective Methods: Self-Reports (I1) and Self-Explanations (I2)**

Is ChatGPT a semantic zombie? Let's ask the system itself. Here are typical excerpts from dialogues with version 3.5 from November 2023:

> Q: "As a language model, you generate any form of text. But do you understand these texts? Can you grasp the meaning of the texts?"
>
> A: "I can generate and understand texts based on patterns and information present in the data on which I was trained. ... I don't have understanding or consciousness in the way humans do. ... My responses are generated by predicting the next word or sequence of words given the input I receive. While I can generate contextually relevant and coherent responses, I lack true comprehension or awareness."

---

[4] An unfortunate expression, it would be more precise to speak of "confabulations", since hallucinations are deceptive perceptions that make no sense for non-modal LLMs. But the term has already caught on.

> Q: "Wouldn't it be possible for you to develop a form of world knowledge or semantic understanding of the world through your training using large amounts of language data?
>
> A: "During my training with large amounts of text data, I have developed a certain amount of knowledge about the world and an understanding of many topics covered in these texts. However ... I have no capacity for self-reflection or thinking for myself. In a sense, you could say that I have an "artificial knowledge of the world" based on the available textual data, but it is not a true understanding or awareness that humans have."

As a first observation: ChatGPT's answers not only contain self-reports but also elements of self-explanations. In this respect, LLMs perform even better than most people. ChatGPT does, however, not separate phenomenality from intentionality. The overall response behavior suggests that the fine-tuning of OpenAI is designed to keep the system away from any suspicion of consciousness or personhood. Recent findings also indicate that prompting is crucial when interacting with LLMs. Prompt engineering has even become a professional skill. Terrence Sejnowski (2023) has argued that LLMs reflect the intelligence of their users. They tune in to their interlocutors. In Sejnowski's words: "In mirroring the interviewer, LLMs may effectively be carrying out a much more sophisticated reverse Turing test, one that tests the intelligence of our prompts and dialog by mirroring it back to us." This analysis is consistent with the Theory-of-Mind-capabilities of GPT-4 already highlighted in the foregoing section 2.1.

Taken together, this shows that introspective methods are not reliable and are subject to at least as much suspicion as E1 methods. On the other hand, the situation in the field of AI does not differ significantly from that in humans. Introspection can also be criticized in psychology with good reasons (cf. Schwitzgebel 2019). With regard to self-explanations, AI systems perform even better than humans, who can generally provide little, if any, information about the cognitive background to their behavior.

## 3 Semantic grounding: applying theories of semantics

After critical examining four out of five methods, we can now turn to E2, the most promising of all methods. Philosophy of mind and philosophy of language have long discussed and developed theories of semantics and mental content. In greatest generality, three important theoretical strands can be distinguished: causal-teleosemantic theories, use theories, and functional role semantics. The first two groups are externalist in nature, the third internalist. Each group allows to distill a general criterion that contributes to the possibility of semantic grounding. This leads to a threefold distinction:

> 1) causal grounding,
> 2) social grounding, and
> 3) functional grounding.

In other words, the semantic capabilities of intelligent agents or systems develop along three dimensions that span a three-dimensional developmental space of semantic grounding (cf. Lyre 2020). This goes hand in hand with the important insight that semantic grounding isn't a yes-no matter, but rather a matter of degree. Intelligent or cognitive agents and systems can be more or less semantically grounded, they can speak or think more or less meaningfully. This should be familiar to us humans: we are born without any semantic grounding and become increasingly more grounded as we grow up. And we can gradually lose this capability, as in the case of mental confusion or dementia. Let us turn to each of the three types of grounding, starting with functional grounding.

## 3.1 Functional grounding

The core idea of functional role semantics and, hence, functional grounding is that meaning is characterized by functional roles. What, for instance, is the meaning of the bishop in chess? It is obviously the piece that moves and captures along diagonals. Chess doesn't exist in space and time, chess pieces do not refer to any worldly items. Each chess piece is given meaning solely by the functional role assigned to it by the rules of the game. An AI system like MuZero (Schrittwieser et al. 2020), a further development of DeepMind's famous Alpha series, which can learn virtually any known board game from scratch through pure self-play (i.e. including the rules of the game itself!) and then master the game at a superhuman level, acquires the meaning of chess in terms of a functional role semantics.

Put another way, MuZero has generated a model of chess by extracting crucial patterns in the self-play training data. These patterns correspond to the functional roles of the pieces in the game and their complex relationships amongst each other. Since chess pieces and their positions do not refer, such functional role semantics is sufficient to play chess meaningfully. The generative model of MuZero has a functional grounding. It is important to emphasize that such grounding is not simply based on MuZero's performance. In 1997, DeepBlue beat then-world chess champion Garry Kasparov to become the first machine to play chess at a superhuman level. This success, however, was largely based on pre-implemented heuristics and brute-force search in the combinatorial tree. In contrast, MuZero's functional grounding is based on a generative model developed via self-learning, the essential feature of any true generative AI.

My rather generic use of the terms functional role semantics and functional grounding is meant to embrace the many flavors discussed in the literature such as computational, inferential, or conceptual role semantics (cf. Block 1998; Cummins 1989, chap. 9; Whiting 2010). My usage is also compatible with inferential semantics as recently discussed in the welcome papers by Søgaard (2022, 2023) and Piantadosi & Hill (2022) in the context of LLMs. Of course the elements of language, words or sentences, have functional roles. Expressions such as "and" and "or" function as logical and computational connectives, and whoever says that "ants are insects" has also tacitly said that "ants are animals", since the second sentence is a conclusion from the first. Sentences or propositions form networks through the inferential relationships in which they stand. Computational and inferential roles are special cases of the general concept of functional roles, and even more generally we can view networks of functional roles as precisely those data structures that capture how single data relate to each other. In this sense, many AI systems, and clearly modern LLMs, extract patterns and structures in data and thus gain a functional grounding.

**3.2 Social grounding**

Functional grounding is a purely system-internal matter, no direct connection between LLMs and the world has yet been taken into account. Social and causal grounding are two principled ways of doing so. And, interestingly enough, social grounding can be viewed as a natural extension of functional grounding. The idea is to extend the concept of functional roles beyond the system boundary. This leads to the functional roles that cognitive agents and systems play in the world through their behavior. Block (1998) has dubbed them "long-arm" functional roles as opposed to the traditional short-arm roles that stop at the system boundary. Long-arm functional roles of linguistic behavior are reminiscent of Wittgensteinian use-theoretic conceptions of meaning, where linguistic meaning is constituted by the functional roles of the speaker's utterances or expressions in the language game. Thus, meaning manifests itself in the use of language and is bound to the existence of a linguistic community. Accordingly, rule following (whether in the sense of grammar or otherwise) is considered by Wittgenstein (1953) as a social practice. We humans have a social grounding precisely to the extent that we are participants in the language games of our language communities.

Such ideas can directly be applied to LLMs. They become participants in our language games precisely to the extent that we increasingly include them in our linguistic practices. This process makes it particularly clear that social grounding, and indeed semantic grounding in general, is gradual in nature. The more useful LLMs are, the more they are integrated into our linguistic practices, and the more they become established as communicative partners, the more they acquire a social grounding. This happens in full analogy to children who gradually grow into their language communities. It is straightforward to assume that the strongest LLMs have already acquired an elementary social grounding comparable to what we find in young children.

Of course, the weak and elementary social grounding of LLMs comprises linguistic behavior only. Lacking any embodiment and interventional skills, the behavior of LLMs is limited to language-based behavior and cannot (yet) possess the full grounding open to causally interactive agents or systems. At this point social and causal grounding overlap. The example of chess from the previous section is illustrative and tricky in this respect. Wittgenstein made it vehemently clear that "playing games" is a practice (and he used the example in a variety of ways to discuss, for instance, family resemblances, language games, or rule following). Following Wittgenstein, MuZero doesn't "play" chess (nor does any other chess program), since current AI systems still lack the social context: the shared and public exercise of games as a behavioral practice. Hence, LLMs possess a functional grounding as well as a weak social grounding limited to language-based behavior.

**3.3 Causal grounding**

Referential semantics regards causal anchoring and reference as the most essential features of semantic grounding. For instance, the extension or reference of singular terms such as proper names is the object denoted by them. Causal theories of reference in the succession of Kripke and Putnam assume the existence of a causal chain that ultimately fixes the reference. Similarly, causal covariance theories of mental content in the succession of Fodor regard causal grounding as their central starting point.

Current LLMs have not yet reached a direct causal grounding (modulo first multimodal approaches and approaches in robotics). However, the models are trained with large amounts of data made available on the Internet. This data has mostly been produced by us humans, i.e. causally grounded beings. This opens the door to a sort of *indirect causal grounding*. Such indirect grounding is functionally established by the stunning fact that modern LLMs develop *world models*, i.e. representations that are structurally similar to (parts of) the world. Put differently: the totality of text and language data is like a huge mirror of the world created by us humans, and modern LLMs are capable of extracting law-like world structures and regularities from the huge amounts of text data from which they learn. Indeed, the breathtaking successes of recent LLMs would be downright mysterious if we did not make the assumption that these systems form grounded world models, even if only in an indirect causal manner.

In fact, an increasing number of recent studies suggest that LLMs possess world knowledge (see examples below). The focus is on proving that the systems develop internal representations whose state spaces have geometries for which semantic relations are crucial. Such *representational geometries* possess structural similarities to the world. Let us consider this in more detail: The n-dimensional state space of a network of n neurons is spanned by the n axes of the activation strength of the individual neurons. A distributed neuronal representation represents a point in this state space (or a vector with the respective individual activations as components). The interesting question is whether and if so which patterns and structures can be found in these activation spaces. Even the simplest classifier networks form activation clusters that follow a semantic interpretation. This is to be expected, since encoder networks are supposed to learn semantically relevant classes. After a network has been trained, these classes show up as clusters in the state space. Crucially, the neighborhood relations of the activation clusters follow semantic proximity or adjacency. Churchland (1989) proposed the concept of a state-space semantics for connectionist systems. The general picture that emerges is that neural state spaces have a semantic interpretation, a representational geometry. This general picture can be found in various approaches: second-order isomorphisms (Shepard & Chipman 1970, Edelman 1998), conceptual spaces (Gärdenfors 2000), or quality spaces for phenomenal content (Clark 2000, Rosenthal 2015, Lyre 2022).

In recent cognitive and computational neuroscience, multivariate pattern analysis methods have been developed to experimentally study representational geometries. Representational similarity analysis (RSA) allows to compare two datasets of different origins using a representational dissimilarity matrix (RDM) (Kriegskorte & Kiewit 2013). The RDM is a measure of negative structural similarity (dissimilarity). Structural similarity is mathematically expressed by homomorphisms, i.e. structure-preserving maps. Structures can be generally understood as sets of objects or relations on which certain relations are imposed. Structural similarity is therefore a second-order similarity, which does not refer to the direct first-order similarity of relations (of the same type), but to the similarity of structurally corresponding relations. In RSA, the datasets to be compared can be of entirely different origins, so the method allows comparisons between otherwise different domains, be it brain activity data (e.g. fMRI or EEG), behavioral data (e.g. reaction times or psychophysical similarity), or data from computational models (e.g. LLMs).

Among the notable studies on LLM world models are the following: Coenen et al. (2019) from Google Brain studied and partially visualized the representational geometry of the Transformer model BERT. Liétard et al. (2021) show that LLMs have some degree of geographic world knowledge. Abdou et al. (2021) show that the structure of the color space, i.e. the quality space of color perception, can be reconstructed from LLM data even though the models themselves do not "perceive" in any way. Furthermore, Patel & Pavlick (2022) show that different conceptual spaces can be generated even by a few training examples and show robustness to isomorphisms such as rotations. Overall, the non-trivial finding is that LLMs are able to extract extensive knowledge about causal and other regularities of the world from vast amounts of textual data through their self-training. In other words, LLMs develop world models by extracting world structure from training data. This gives the systems an indirect causal grounding. The grounding is indirect because the causal chain – the causal contact of the systems with the world – does not run through direct perceptual or even sensorimotor processes, but is mediated by human language data, i.e. mediated by us directly causally grounded, human speakers.

To summarize: LLMs not only perform word statistics, but also develop world models. Self-learning in LLMs is based on statistical learning methods that reveal not only syntactic but also semantic regularities and inferential structures. This already corresponds to a functional grounding, but it also corresponds to an indirect causal grounding, as these extracted structures are world models. They are (partially) isomorphic to structures of the world, as witnessed by representative geometries.

## 4 Concluding remarks

In this paper, I started with the observation that we have recently witnessed a generative turn in AI. The generative turn lies at the heart of the deep learning revolution, since generative models, including LLMs, are key for self-supervised learning. I then proposed a five-fold methodology to tackle the semantic grounding problem of LLMs. It turned out that the most promising way is to apply core assumptions of theories of meaning in philosophy of mind and language to LLMs. We found that LLMs are neither stochastic parrots nor semantic zombies, but must rather be seen as (i) reasonably functionally grounded, (ii) weakly socially grounded, and (iii) indirectly causally grounded. A strong argument for grounding is that LLMs form world models, and the evidence for this is that the representational geometry of these models follows semantic similarities.

I would like to finish with two points: a short remark, first, on where the LLM world models reside; and, second, a reflection on the prospects of a more direct causal grounding. As pointed out in section 1.2, the transformer training is based on a two-step process. While the first step consists of self-learning and leads to a foundation model, the second step comprises fine-tuning by RLHF. It is mainly the first step, where LLMs acquire a crucial amount of world knowledge in terms of law-like regularities that the systems find and extract from the massive amounts of text data. Hence, through the process of generative self-learning, LLMs form world models encoded as representational geometries. Such world models therfore reside in the parameters (connection strengths) of the foundation models. Moreover, the process of RLHF endows LLMs with a first kind of social grounding.

Now to the second point: What amounts to direct causal grounding? Which conditions have to be fulfilled? I claim that, here again, gradual criteria are to be expected. Consider a thought experiment: Papageno has grown up in a stimulus-poor room lacking any direct causal contact with the world, but he has unlimited written and spoken media at his disposal, such as newspapers, books, radio and telephone (the written media are purely written media, i.e. they do not contain any graphics or pictures). Is Papageno a semantic zombie?

Papageno's situation is essentially the same as that of current LLMs, he has an indirect causal grounding. A natural next step is to endow him with sensory possibilities. What, however, does "sensory" precisely mean? In the case of visual sensation: would the media feed of graphics and images be sufficient? Or, is video or TV necessary? Prima facie, a media-mediated world isn't equivalent to a direct, immediate perception of the world. Any media-mediated information could in principle be fake. Papageno v.2 is no better off than a Putnam-style brain-in-a-vat. His world models are threatened by a principled skepticism. So maybe direct causal grounding is provided by direct perception only. Call this Papageno v.3: he has perceptual, but no motor access to the world. Ultimately, he is no better off than a complete locked-in patient. In a further step, therefore, Papageno v.4 should be fully embodied and endowed with active motor skills.[5]

Of course, this is nothing more than a rough sequence of significant Papageno versions. But whatever the main versions are, the story clearly shows that direct causal grounding, like semantic grounding in general, turns out to be a gradual affair. There is a gradual path from indirectly to rather directly grounded AI systems. And clearly, AI technology is already treading this gradual path with vigor. The development of multimodal LLMs linked to robotics and embodied cognition is in full swing, and innovative systems will not be long in coming. For example: The recent GPT-4V (with vision) allows GPT-4 to support image inputs as much as text. Google has just presented PaLM-E, an embodied LLM that integrates visual input and speech for robot control (Driess et al. 2023). And Google DeepMind has started the era of its multimodal Gemini models that can generalize and seamlessly understand and process text, audio, images and video (Gemini Team 2023).

---

[5] Combining ideas from embodiment, enactivism and predictive processing, Pezzulo et al (2024) argue for active inference as being the decisive factor for semantic grounding.

With the advent of transformer-based LLMs, we have truly entered a new era in AI. This is the most visible element of the generative turn. Current LLMs already show an elementary grounding, and as the development advances at great speed, we will very soon see the transition from indirect to direct causal grounding. Full-blown semantically grounded AI systems, meaningfully talking and thinking machines, are only a matter of time.